\documentclass[a4paper,UKenglish,cleveref, autoref, thm-restate]{lipics-v2021}

\hideLIPIcs  

\title{CP-Model-Zoo: A Natural Language Query System for Constraint Programming Models\footnote{presented at \emph{LLMs meet Constraint Solving} Workshop at CP2025 in Glasgow}} 

\titlerunning{CP-Model-Zoo: A Natural Language Query System for CP Models} 

\author{Augustin {Crespin}}{UCLouvain, Belgium}{augustin.crespin@student.uclouvain.be}{https://orcid.org/0009-0007-9238-8063}{}

\author{Ioannis {Kostis}}{ICTEAM, UCLouvain, Belgium}{ioannis.kostis@uclouvain.be}{https://orcid.org/0000-0002-1875-8587}{}

\author{Hélène {Verhaeghe}}{ICTEAM, UCLouvain, Belgium}{helene.verhaeghe@uclouvain.be}{https://orcid.org/0000-0003-0233-4656}{}

\author{Pierre {Schaus}}{ICTEAM, UCLouvain, Belgium}{pierre.schaus@uclouvain.be}{https://orcid.org/0000-0002-3153-8941}{}

\authorrunning{A. Crespin et al.} 

\Copyright{Augustin Crespin, Ioannis Kostis, Hélène Verhaeghe and Pierre Schaus} 

\ccsdesc[500]{Computing methodologies~Natural language generation}
\ccsdesc[500]{Theory of computation~Constraint and logic programming}
\ccsdesc[500]{Information systems~Recommender systems}

\keywords{Constraint Identification, Large Language Models, Constraint Programming,
Natural Language Processing, Information Retrieval, NLP4CP} 

\supplementdetails[linktext={https://github.com/crespina/LLM4CP}, cite={}, subcategory={}, swhid={Software Heritage Identifier}]{Software (Source Code)}{https://github.com/crespina/LLM4CP} 

\nolinenumbers 

\EventEditors{John Q. Open and Joan R. Access}
\EventNoEds{2}
\EventLongTitle{42nd Conference on Very Important Topics (CVIT 2016)}
\EventShortTitle{CVIT 2016}
\EventAcronym{CVIT}
\EventYear{2016}
\EventDate{December 24--27, 2016}
\EventLocation{Little Whinging, United Kingdom}
\EventLogo{}
\SeriesVolume{42}
\ArticleNo{23}
\usepackage{parskip}
\usepackage{makecell}
\usepackage{amsmath}
\usepackage{appendix}
\usepackage{hyperref}

\usepackage{listings}
\begin{document}

\maketitle

\begin{abstract}
Constraint Programming and its high-level modeling languages have long been recognized for their potential to achieve the holy grail of problem-solving.
However, the complexity of modeling languages, the large number of global constraints, and the art of creating good models have often hindered non-experts from choosing CP to solve their combinatorial problems.
While generating an expert-level model from a natural-language description of a problem would be the dream, we are not yet there.
We propose a tutoring system called \emph{CP-Model-Zoo}, exploiting expert-written models accumulated through the years.
CP-Model-Zoo retrieves the closest source code model from a database based on a user's natural language description of a combinatorial problem. 
It ensures that expert-validated models are presented to the user while eliminating the need for human data labeling. 
Our experiments show excellent accuracy in retrieving the correct model based on a user-input description of a problem simulated with different levels of expertise.
\end{abstract}

\section{Introduction}
\label{sec:intro}

Constraint Programming (CP) and its high-level modeling languages have long been recognized for their potential to achieve the holy grail of problem-solving: "\emph{The user states the problem, the computer solves it}" \cite{freuder1997pursuit}. However, finding the right model is not easy.

First, the learning complexity of pure CP modeling languages (like MiniZinc \cite{mnzc}, XCSP3 \cite{boussemart2016xcsp3}, or Essence \cite{frisch2008ssence}) can pose a challenge even for computer scientists.
Second, even when the API is suggested in a mainstream language like CPMPy \cite{guns2019increasing},
modeling in CP is considered an art with many good practices or intuitions that are difficult to formalize.
As explained in \cite{modelling}, one can use techniques such as breaking symmetries, redundant constraint, using multiple viewpoints with channeling constraints, etc.
It is well known among CP experts that modeling with binary variables, as in MIP or SAT, generally leads to inefficient models for CP. Instead, efficient CP models usually use global constraints (e.g., \texttt{element} \cite{element}, \texttt{table} \cite{str2},...), offering a stronger and more efficient pruning.
Choosing the proper one and their efficiency among many of them\footnote{The catalog of global constraints \cite{catalog} count more than 400 constraints.} is also non-trivial.

With the recent progress in AI and large language models, we can dream of a tool that generates expert-level CP models from a natural language description of a problem to target a CP system. The reality is that we are not yet there and may not be ready to achieve this soon for entirely new problems.
As an example, we have tried feeding this toy puzzle description to state-of-the-art large language models (LLM): ChatGPT-4o, Gemini 2.5 Pro, and Mistral Lechat Pro:

\emph{You are given 13 words: BUOY, CAVE, CELT, FLUB, FORK, HEMP, JUDY, JUNK, LIMN, QUIP, SWAG, VISA, and WISH. These words collectively contain 24 unique letters. The goal is to assign each of these 24 letters to one of the six faces of four different cubes such that each letter appears on exactly one face of one cube. The assignment must ensure that for each word, its four letters are distributed across the four different cubes. Write a MiniZinc model for solving this problem.}

\begin{figure}[ht]
    \centering
\begin{minipage}{0.47\textwidth}
\begin{lstlisting}
include "alldifferent.mzn";

% The 24 unique letters found in the given words
set of int: LETTERS = 1..24;

% The 13 words given
array[1..13, 1..4] of int: WORDS = [
    | 1,  2,  3,  4   % BUOY
    | 5,  6,  7,  8   % CAVE
    | 9,  10, 11, 12  % CELT
    | 13, 14, 15, 16  % FLUB
    | 17, 18, 19, 20  % FORK
    | 21, 22, 23, 24  % HEMP
    | 3,  6,  12, 14  % JUDY
    | 10, 13, 19, 21  % JUNK
    | 4,  7,  9,  11  % LIMN
    | 8,  15, 16, 22  % QUIP
    | 5,  17, 20, 23  % SWAG
    | 1,  18, 24, 11  % VISA
    | 2,  9, 12, 16   % WISH
];

\end{lstlisting}
\end{minipage}\hfill
\begin{minipage}{0.47\textwidth}
\begin{lstlisting}
% Assign each letter to one of the four cubes
array[LETTERS] of 1..4: cube;

constraint alldifferent([cube[i] | i in LETTERS]);

% Ensure that each word's letters are on different cubes
constraint forall(w in 1..13)(
    cube[WORDS[w,1]] != cube[WORDS[w,2]] /\
    cube[WORDS[w,1]] != cube[WORDS[w,3]] /\
    cube[WORDS[w,1]] != cube[WORDS[w,4]] /\
    cube[WORDS[w,2]] != cube[WORDS[w,3]] /\
    cube[WORDS[w,2]] != cube[WORDS[w,4]] /\
    cube[WORDS[w,3]] != cube[WORDS[w,4]]
);
\end{lstlisting}
\end{minipage}
    \caption{Code produced by ChatGPT for the cube problem}
    \label{apx:cube}
\end{figure}

The generated output for ChatGPT is given in Fig.~\ref{apx:cube} and is representative of the ones generated by concurrent services. This code has some serious issues:
\begin{itemize}
\item There are syntactical errors. It also didn't seem to understand the concept of decision variables and introduced plain sets instead.
\item It introduced inconsistent data. Different integers were used to encode the same letters.
\item It correctly tried to assign the letters of the same word to different cubes but used a decomposition instead of a global \texttt{alldifferent} constraint, which is not the best model an expert can imagine.
\item The model does not enforce the cardinality of 6 faces per cube. This is a missing constraint. An expert would model it with a global cardinality constraint.
\end{itemize}
As CP experts, we tried to fix the model, but it took us more time and energy than starting from scratch, mainly because the bugs were not introduced incrementally.
Learning quickly to reach a sufficient level of expertise in CP is, in our opinion, the safest approach to write good models. A CS professor arrived at a similar conclusion after experimenting with a programming task solved by students using LLM and without it \cite{impact_ai_cs_education}. One effective way to learn is by studying existing models that solve problems similar to the one at hand. 

In this paper, we propose a pragmatic approach relying upon the wealth of high-quality models accumulated over the years by the community through solver competitions \cite{stuckey2014minizinc,audemard2024proceedings}, CSPLib \cite{gent1999csplib}, and system tutorials \cite{hakank_github}. The proposed solution retrieves the closest source code model from a database of models based on a user's natural language description of a combinatorial optimization problem. While utilizing recent advancements in LLM technology, this approach is designed to avoid hallucinations, ensuring that only expert-validated models are presented to the user. It also eliminates the need for human intervention in data labeling.
The system, called CP-Model-Zoo\footnote{Name inspired by the Scheduling Zoo (\url{http://schedulingzoo.lip6.fr/}), a tool to retrieve scientific articles based on the main characteristics of a scheduling problem.}, is both fast and incremental. Adding a new model only requires generating a natural language description using an LLM and then computing a language model embedding in a vector space. Retrieving the most relevant model involves computing an embedding of the query and finding the model in the database with the closest similarity in the embedding space, as measured by cosine similarity.

Through experiments, we demonstrate that our system achieves excellent accuracy in retrieving the correct model based on user input descriptions simulated at different levels of expertise. This paper details our methodology and introduces CP-Model-Zoo, an open-source tool available\footnote{\url{https://cp-model-zoo.info.ucl.ac.be/}} to the CP community and its potential users. We believe this tool offers a practical contribution to enhancing the learning and application of CP for a broader audience.

The paper is organized as follows. Section \ref{sec:related} presents the related work, providing an overview of existing research in the field. Section \ref{sec:method} outlines the methodology to rank and select the model starting from the natural language query. 
Section \ref{sec:exp} conducts experiments to validate the tool's capacity to retrieve relevant models from text descriptions of varying levels of expertise.
Section \ref{sec:web} describes the tool's user interface. Finally, Section \ref{sec:concl} concludes the paper by summarizing the key findings and discussing future directions.

\section{Related work}
\label{sec:related}

Different approaches have been tried to close the gap between CP formalism and natural language descriptions of CP problems.

First, the \textit{NL4Opt} competition at \textit{NeurIPS} in 2023 formulated the challenge of using \textit{NLP} methods for extracting the meaning and formulation of an optimization problem based on its text description \cite{ramamonjison_nl4opt_2023}. The top-performing approaches in the competition \cite{gangwar_highlighting_2023, ning_novel_2023, jang_tag_2022, he_linear_2022} mainly relied on fine-tuned BERT \cite{devlin_bert_2019} and BART \cite{lewis_bart_2020} architectures.

Hahn et al. \cite{hahn_formal_2022} fine-tuned the base version of the open-source language model T5 \cite{raffel_exploring_2023} to transform natural language descriptions into regex or other formal descriptions. Karia et al. \cite{karia_can_nodate} used LLMs to transform SAT formulas into natural language and vice-versa with a 95\% 

Tsouros et al. \cite{tsouros_holy_2023} tried to enhance the capacities of plain (not fine-tuned) LLMs with different techniques to make them CP modelers in CPMpy. The pipeline comprises a Named Entity Recognition to identify essential features of the problem, a relationship identifier between these named entities, and then a translation into CPMpy code. The system automatically corrects the errors in a loop, allowing users to refine the model iteratively. For modeling, a few LLM tricks are used, such as system prompting, chain or tree of thoughts, and plan-and-solve. The proposed method demonstrates promising results on selected examples; however, its generalizability to real-world datasets remains unverified. 

Hotz et al. \cite{hotz_exploiting_2024} made an LLM find solutions for the most well-known CP problems by iterating with the PyChoco API\footnote{\url{https://pychoco.readthedocs.io/en/latest/}} 
to find a syntactically correct solution and then a semantically correct one. The results were mitigated. Solutions were found for well-known problems (N-Queens, Map Coloring, TSP), but the system often hallucinated PyChoco functions, and the chats were unstable. Szeider used a similar idea in his MCP-Solver \cite{szeider_mcp-solver_2024}. The client enables an LLM to interact with constraint models through natural language, while the server manages these interactions and translates them into MiniZinc operations using its Python API. A system prompt guides the LLM’s interaction. Likewise, solutions were found for famous problems, but no comparison with a plain LLM as a baseline was provided.

Michailidis et al. \cite{michailidis_constraint_2024} used LLMs as CP solvers and modelers. Without surprises, using plain LLMs as solvers only yielded an accuracy of 11.46\% on the \textit{NL4Opt} dataset, showing clearly the need for a more specialized system to handle this task. For modeling,  two techniques were used, i.e., blueprint model generation and in-context learning, with dynamically chosen examples. They were able to attain an 87.54 \% solution accuracy on the \textit{NL4Opt} dataset and 76.00\% accuracy on a Logic Grid Puzzles dataset.

Let's note that Voboril et al. \cite{voboril_realtime_2024} proposed StreamLLM, a system capable of generating streamliners (i.e., constraints added to a model to reduce the search space) in real time using LLMs. However, it relies heavily on the fact that a functional model already exists for the problem.

A more interactive approach, specifically a decision support for the meeting scheduling problem, was used by Lawless et al. \cite{lawless_i_2024}. An LLM is used in the Constraint Management component. It is prompted to select an action to take (add/delete a constraint, generate a suggestion, etc.). If a new constraint is added, two LLMs are used: an information checker and a coder in Python. It achieved around 80\% accuracy for the information checking task and generated code that compiles correctly over 95\% of the time and has high precision and recall (over 90\%). 

Similarly, Chen et al. \cite{chen_optichat_2025} presented OptiChat, a natural language dialogue system intended to assist users in interpreting optimization model formulation, diagnosing infeasibility, doing a sensitivity analysis,
retrieving data and evaluating modifications. Users upload a Pyomo-based optimization model \cite{pyomo}, and can then interact with the model through natural language queries, asking about feasibility, sensitivity, etc.
The evaluation is conducted on a dataset of Pyomo-based models with a success rate of above 80\% for a majority of query types, with particularly strong performance in diagnosing infeasibility (87.2\%) and sensitivity analysis (94.4\%), while significantly reducing response time compared to human experts. 

Contrary to related work, we do not use LLMs to directly model or solve the problem. Instead, we propose a system that takes input from a user's description of a combinatorial problem and outputs the closest CP model from a database of models. Therefore, our system is closer in philosophy to an Information Retrieval (IR) system.

\section{Methodology and problem description}
\label{sec:method}

\begin{figure}[ht]
    \centering
\begin{subfigure}{0.47\textwidth}
\centering
\begin{lstlisting}
    You are given one or more MiniZinc models that represent a 
    single classical constraint programming problem. Your task is to
    read the code and explain what the problem is about using very 
    simple language. If there are several models for the same 
    problem, do not explain each one separately. Instead, focus on 
    explaining the overall problem. Assume the reader does not have 
    much background in programming or mathematics. In your answer 
    please explain: 
    The name of the problem.
    What the problem is about in everyday terms.
    What the main variables are and what they mean, using plain language.
    What the basic restrictions or rules of the problem are, explained simply.
    What the goal of the problem is (for example, what you want to minimize or maximize).
    In your answer, do not include any introductory phrases (such as 
    'Here is the explanation of the problem')
    Here is the source code:
    --------------
    {source_code}
    --------------
\end{lstlisting}
    \caption{Novice Level - D1}
    \label{apx:novice}
\end{subfigure}
\begin{subfigure}{0.47\textwidth}
\centering
\begin{lstlisting}
    You are experienced in constraint programming and familiar with 
    MiniZinc. You are provided with one or more MiniZinc models 
    representing a classic constraint programming problem. Your task 
    is to identify the problem and explain it in clear, 
    intermediate-level language. Assume the reader has some 
    technical background but is not an expert. If there are several 
    models for the same problem, do not explain each one separately. 
    Instead, focus on explaining the overall problem. In your answer 
    please explain:
    The name of the problem.
    A concise description of what the problem is about.
    An explanation of the main decision variables and what they represent.
    A description of the key constraints in plain language 
    (avoid heavy mathematical notation).
    An explanation of the problem's objective (what is being minimized or maximized).
    In your answer, do not include any introductory phrases (such as 
    'Here is the explanation of the problem')
    Here is the source code of the model(s):
    --------------
    {source_code}
    --------------
\end{lstlisting}
    \caption{Intermediate Level - D2}
    \label{apx:intermediate}
\end{subfigure}

\begin{subfigure}{0.6\textwidth}
\centering
\begin{lstlisting}
    You are an expert in high-level constraint modeling and solving discrete 
    optimization problems. In particular, you know MiniZinc. You are provided 
    with one or several MiniZinc models that represents a single classical 
    problem in constraint programming. Your task is to identify what is the 
    problem modeled and give a complete description of the problem to the user.
    If there are several models for the same problem, do not explain each one 
    separately. Instead, focus on explaining the overall problem.
    This is the source code of the model(s):
    --------------
    {source_code}
    --------------
    In your answer please explain:
    name: The name of the problem
    description: A description of the problem in English
    variables: A string containing the list of all the decision variables in 
    mathematical notation, followed by an explanation of what they are in English
    constraints: A string containing the list of all the constraints in 
    mathematical notation, followed by an explanation of what they are in English
    objective: The objective of the problem (minimize or maximize what value)
    In your answer, do not include any introductory phrases 
    (such as 'Here is the explanation of the problem')
\end{lstlisting}
    \caption{Expert Level - D3}
    \label{apx:expert}
\end{subfigure}
    
    \caption{Prompts to Generate Descriptions}
    \label{apx:prompts}
\end{figure}

Let $\mathcal{D} = \{s_1, s_2, \dots, s_N\}$ be a database of strings, where each $s_i$ represents a source code implementation of a combinatorial optimization problem, possibly appended with an enriched textual description of the problem that is LLM-generated from the source code.

Given a query string $q$ (a natural language problem description entered by the user), our goal is to retrieve the top-$k$ most semantically similar strings from $\mathcal{D}$ to the query string, therefore returning the most similar source-code implementation to help the user model their problem. 

We use a pre-trained text embedding model to project strings into a shared embedding space. Let:
\begin{itemize}  
    \item $\phi: \mathcal{S} \rightarrow \mathbb{R}^d$ denote the embedding function, where $\mathcal{S}$ is the string space and $d$ is the embedding dimension.
    \item $\mathbf{e}_i = \phi(s_i)$ be the precomputed embedding vector of database string $s_i$.  
    \item $\mathbf{e}_q = \phi(q)$ be the embedding vector of the query, computed at inference time.  
\end{itemize}  
Semantic similarity is measured via cosine distance in the embedding space:  
$$
\text{sim}(q, s_i) = \cos(\mathbf{e}_q, \mathbf{e}_i) = \frac{\mathbf{e}_q \cdot \mathbf{e}_i}{\|\mathbf{e}_q\| \|\mathbf{e}_i\|}
$$ 
The top-$k$ retrieved items are the $s_i \in \mathcal{D}$ with the highest similarity scores.

To construct $\mathcal{D}$, various strategies are possible:
\begin{enumerate}  
    \item \textbf{Code-only}: Raw source code strings.  
    \item \textbf{LLM-augmented}: Source Code + Synthetic Descriptions of varying expertise levels:  
    \begin{itemize}  
        \item \textit{\textbf{Novice}}: Very simple explanations in non specific terms.

        The prompt used to generate such descriptions can be found in Fig.~\ref{apx:novice}, while a sample \textit{Novice}-level description for the knapsack problem is included in Fig.~\ref{apx:sample_cl1}.
                
        \item \textit{\textbf{Intermediate}}: Domain-specific terminology.
        
        Similarly, the prompt used to generate such descriptions can be found in Fig.~\ref{apx:intermediate}. In contrast, a sample \textit{Intermediate}-level description for the knapsack problem is included in Fig.~\ref{apx:sample_cl2}.
                
        \item \textit{\textbf{Expert}}: Formal mathematical formulations.
        
        The prompt to be provided to an LLM to generate this level of descriptions can be found in Fig.~\ref{apx:expert}, while a sample \textit{Expert}-level description for the knapsack problem is included in Fig.~\ref{apx:sample_cl3}.
                
    \end{itemize}  
\end{enumerate}

\begin{figure}[ht]
    \centering
\begin{subfigure}{\textwidth}
\centering
\begin{lstlisting}
**Knapsack Problem**

This problem is about packing objects of different sizes and values into a bag with a limited capacity. Imagine you're going on a trip and you have 
a small bag that can only hold so much weight. You have several items you want to bring, each with a different weight and importance to you. You 
want to pack the most important items that fit in the bag without exceeding its weight limit.

**Main variables:**

* `n`: the number of objects
* `capacity`: the maximum weight the bag can hold
* `profit` (or `value`): how important each object is to you
* `size`: how much each object weighs
* `x`: which objects to pack in the bag (how many of each object to take)

**Basic restrictions or rules:**

* You can't pack more objects than you have.
* The total weight of the packed objects can't exceed the bag's capacity.
* You can't pack a fraction of an object (you either take it or leave it).

**Goal:**

The goal is to pack the objects that give you the most importance (or value) while staying within the bag's weight limit.
\end{lstlisting}
    \caption{Novice Level}
    \label{apx:sample_cl1}
\end{subfigure}

\begin{subfigure}{0.47\textwidth}
\centering
\begin{lstlisting}
The problem is known as the 0/1 Knapsack Problem.

This problem is about selecting a subset of items to include in a 
knapsack of limited capacity, with the goal of maximizing the total
value of the selected items. Each item has a size and a profit, and 
the knapsack has a limited capacity.

The main decision variables are the `x` variables, which represent 
whether each item is included in the knapsack (1) or not (0).

The key constraints are:

* Each item can only be included once (or not at all), which is ensured by the `x` variables being binary.
* The total size of the included items must not exceed the knapsack's capacity.

The objective is to maximize the total profit of the included items.

In other words, the problem is to find the optimal selection of 
items to put in the knapsack, given the capacity constraint, to 
maximize the total value of the selected items.
\end{lstlisting}
    \caption{Intermediate Level}
    \label{apx:sample_cl2}
\end{subfigure}
\begin{subfigure}{0.47\textwidth}
\centering
\begin{lstlisting}
name: 0/1 Knapsack Problem

description: The 0/1 Knapsack Problem is a classic problem in 
combinatorial optimization. Given a set of objects, each with a 
weight and a value, determine the subset of objects to include in a 
knapsack of limited capacity to maximize the total value while not 
exceeding the knapsack's capacity.

variables: x (x[i] for i in OBJ or x[i] for i in 1..n)

The decision variables x represent the number of each object to 
include in the knapsack. In some models, x is a binary variable (0 
or 1) indicating whether to include the object or not, while in 
others, x is an integer variable representing the quantity of each 
object to include.

constraints: x[i] >= 0, sum(i in OBJ)(size[i] * x[i]) <= capacity

The constraints ensure that the number of each object included is 
non-negative and that the total weight of the selected objects does 
not exceed the knapsack's capacity.

objective: maximize sum(i in OBJ)(profit[i] * x[i])

The objective is to maximize the total value of the objects included 
in the knapsack.
\end{lstlisting}
    \caption{Expert Level}
    \label{apx:sample_cl3}
\end{subfigure}
    \caption{Description Samples for the Knapsack Problem}
    \label{apx:sample}
\end{figure}

At inference time, only $\mathbf{e}_q$ requires computation, since the database embeddings $\{\mathbf{e}_i\}_{i=1}^N$ are precomputed. 
Retrieval reduces to:  
$$
\argmax_{s_i \in \mathcal{D}}^k sim(\mathbf{e}_q, \mathbf{e}_i))
$$
where $k \ll N$ is the number of models to return to the user.

Adding a new model to this framework is incremental and only requires 1) using an LLM to generate synthetic descriptions and 2) computing the corresponding embedding and storing it.


\section{Experiments}
\label{sec:exp}

Our experiments evaluate the capacity of our approach to rank highly relevant models. We designed an experimental setting that, starting from the set of models, does not require human input for the query while ensuring that the ground truth model related to the query is present.

All experiments were implemented\footnote{\url{https://github.com/crespina/LLM4CP}} using Python 3.10. \textbf{LlamaIndex} \cite{llamaindex} was used for orchestration, while \texttt{Llama3-70b-8192} served as the LLM \cite{llama3modelcard} to generate the synthetic descriptions. The LLM is hosted and served through an API provided by \textbf{Groq}\footnote{\url{https://console.groq.com/}} to ensure fast text generation. The embedding\footnote{\url{https://huggingface.co/Alibaba-NLP/gte-modernbert-base}} model is built upon the latest modernBERT pre-trained encoder-only foundation model \cite{modernbert}, with an embedding vector size of $d=768$ \cite{zhang2024mgte}.

The experiments rely on two datasets origins: the \textbf{MiniZinc examples}\footnote{\url{https://github.com/MiniZinc/minizinc-examples/tree/master}}, and the \textbf{CSPLib}\footnote{\url{https://github.com/csplib/csplib}} problems, which included both a MinZinc model and their corresponding human-written natural language descriptions. Among the 67 problems included in the dataset, 36 have a CSPLib description. We used only the Minizinc files as our database. The CSPLib descriptions were used as a validation set to verify whether our system could retrieve the correct Minizinc file (i.e., the associated Minizinc file on the CSPLib website was the truth value).

As a metric for evaluation, we use Mean Reciprocal Rank (MRR) \cite{mrr}, defined as: 
$\mathrm{MRR} = \frac{1}{|Q|}\sum_{i=1}^{|Q|} \frac{1}{rank_i}$ where $Q$ represent a set of queries and $rank_i$ refers to the rank position of the correct source-code model for the \textit{i}-th query. An $\mathrm{MRR}$ of 1 means that the corresponding source-code model was ranked first for each query.

Different embeddings were created for each model, denoted as SC+\{DX\}.
The first embedding, SC, is computed solely from the MiniZinc source code. 
The other embeddings SC+\{DX\} are computed from the source code plus the LLM-generated text description with three different levels of expertise: Novice(1), Intermediate(2) and Expert(3). The prompts used to configure and generate them are displayed in Appendix \ref{apx:prompts}.

The experiment follows a leave-one-out approach between the query and each configuration for the embedding input.
The results in Table \ref{tab:loo} report the MRR value computed on the ranking of the top-5 models retrieved out of the 67 models in total, which served as candidate models for selection, using different input of embeddings. 
These inputs included the source code alone, as well as the source code combined with LLM-generated descriptions of the problem at different levels of expertise. 
We tested four categories of queries. The first three correspond to the three levels of LLM-generated descriptions (67 queries for each level). They were compared to various embeddings containing a combination of the other descriptions (leave-one-out). The truth value here is the source code used to generate the description.
The fourth one consists of the CSPLib human-written descriptions (36 queries), for which we have a truth value in the form of the Minizinc files associated with the website descriptions that were included in the source code database.


\begin{table}[h]
    \centering
    \caption{MRR values computed on the CP models, which served as candidate models for selection, using different sources of embeddings.}
    \label{tab:loo}
    \begin{tabular}{c c cccccccc}
        & & \multicolumn{7}{c} {\textbf{Source of Embedding}} \\ 
        \hline
        & & \makecell[c]{SC} & \makecell[c]{\vspace{-0.1cm}SC +\\ D1} & \makecell[c]{\vspace{-0.1cm}SC +\\ D2} & \makecell[c]{\vspace{-0.1cm}SC +\\ D3} & \makecell[c]{\vspace{-0.1cm}SC +\\ D1\&2} & \makecell[c]{\vspace{-0.1cm}SC +\\ D1\&3} & \makecell[c]{\vspace{-0.1cm}SC +\\ D2\&3} & \makecell[c]{\vspace{-0.1cm}SC +\\ D1\&2\&3} \\[0.3em]
        \hline
        \multirow{3}{*}{\rotatebox{90}{\textbf{Query\hspace{0.45cm}}}} 
        & \text{D1} & 0.9851 & - & \textbf{1.0} & 0.9925 & - & - & \textbf{1.0} & - \\
        & \text{D2} & 0.9254 & 0.9378 & - & \textbf{0.9577} & - & 0.9478 & - & - \\
        & \text{D3} & 0.9527 & 0.9639 & \textbf{0.9925} & - & 0.9776 & - & - & - \\
        & \text{CSPLib} & 0.8736 & 0.8898 & \textbf{0.9051} & 0.8676 & 0.8796 & 0.8704 & 0.8829 & 0.8634 \\[0.3em]
        \hline
    \end{tabular}
\end{table}

The embeddings solely relying on MiniZinc source codes consistently perform worse than the others, supporting the hypothesis that attaching generated textual descriptions to the source code is beneficial.
Notably, when \textit{Novice}-level descriptions are used as queries, the embeddings containing both source codes and \textit{Intermediate}-level descriptions, as well as the embeddings containing source code along with both \textit{Intermediate}- and \textit{Expert}-level descriptions, achieve an MRR of 1.0. This means that, for all problems, the corresponding model is correctly identified and ranked first every time. Furthermore, the scores remain consistently above $0.9$, often exceeding $0.95$. 
The quality of the results when the embeddings are based solely on the source code is noteworthy. While expecting lower MRR values in this configuration, we believe this can be partially attributed to some comments in the source code and the choice of variable identifiers, which add important semantic information that could aid the retrieval task.

\begin{figure}[ht]
    \centering
\begin{subfigure}{0.44\textwidth}
\centering
\begin{lstlisting}
Question: Dart competitions in pubs often have prizes for the 
first, second, third and fourth best throwers. What is the 
order of play for a darts tournament involving n throwers that:
- identifies the best k throwers (and their order);
- has the least number of matches thrown;
- has the least number of games on each of m dartboards (since tournaments typically run in parallel);
- and is the most exciting (that is, prize winners should be discovered at the last possible minute so that everyone stays drinking beer until the end of the tournament).

Ranking:
Rank 1:
Problem Name: table_seating (Similarity: 0.6665750944113722)   
Rank 2:
Problem Name: TTPPV (Similarity: 0.6545532118345632)      
Rank 3:
Problem Name: golfers(Similarity: 0.6530544088881645)      
Rank 4:
Problem Name: cluster (Similarity: 0.6466407509100225)   
Rank 5:
Problem Name: doublechannel (Similarity: 0.6451558866521051)      
\end{lstlisting}
    \caption{Darts tournament (CSPLib Prob 20)}
    \label{apx:sample_cl3}
\end{subfigure}
\begin{subfigure}{0.55\textwidth}
\centering
\begin{lstlisting}
Question: In the railway domain, the action of directing the traffic in 
accordance with an established timetable is managed by a software. 
However, in case of real time perturbations, the initial schedule may become 
infeasible or suboptimal. Subsequent decisions must then be taken manually by an
operator in a very limited time in order to reschedule the traffic and reduce 
the consequence of the disturbances. They can for instance modify the departure 
time of a train or redirect it to another route. Unfortunately, this kind of 
hazardous decisions can have an unpredicted negative snowball effect on the 
delay of subsequent trains. Basically, the goal is to schedule adequately 
trains in order to bring them to their destination. The decision is to chose, 
for each train, which route it will follow and at what time.

Ranking:
Rank 1:
Problem Name: bus_scheduling (Similarity: 0.670888769314417)  
Rank 2:
Problem Name: doublechannel (Similarity: 0.649158749666409)   
Rank 3:
Problem Name: constrained_connected (Similarity: 0.6463022454817552)   
Rank 4:
Problem Name: nurses (Similarity: 0.6382525497004593)   
Rank 5:
Problem Name: rcpsp (Similarity: 0.6369954887069154)   
\end{lstlisting}
    \caption{Train Traffic Rescheduling (CSPLib Prob 78)}
    \label{apx:sample_cl3}
\end{subfigure}

\begin{subfigure}{\textwidth}
\centering
\begin{lstlisting}
Question: Specification - Type 1
Steel is produced by casting molten iron into slabs. A steel mill can produce a finite number, sigma, of slab sizes. An order has two properties, a 
colour corresponding to the route required through the steel mill and a weight. Given d input orders, the problem is to assign the orders to slabs, 
the number and size of which are also to be determined, such that the total weight of steel produced is minimised. This assignment is subject to 
two further constraints:
Capacity constraints: The total weight of orders assigned to a slab cannot exceed the slab capacity.
Colour constraints: Each slab can contain at most p of k total colours (p is usually 2).
The colour constraints arise because it is expensive to cut up slabs in order to send them to different parts of the mill.
The above description is a simplification of a real industrial problem. For example, the problem may also include inventory matching, where surplus 
stock can be used to fulfil some of the orders.
Specification - Type 2
The Type 1 specification does not constrain the number of slabs used. The Type 2 specification extends the objective to require further that the 
number of slabs used to accommodate the minimal weight of steel is also minimised.

Ranking:
Rank 1:
Problem Name: carpet_cutting (Similarity: 0.6201213094705699)
Rank 2:
Problem Name: vessel-loading (Similarity: 0.6084444196269146)
Rank 3:
Problem Name: toomany (Similarity: 0.6037839776563689)
Rank 4:
Problem Name: car (Similarity: 0.6034512332968589)
Rank 5:
Problem Name: sonet_problem (Similarity: 0.6028041619673954)
\end{lstlisting}
    \caption{Steel Mill Slab Design (CSPLib Prob 38)}
    \label{apx:sample_cl3}
\end{subfigure}
    
    \caption{Problems not in the database}
    \label{anxD:other}
\end{figure}

Interestingly, despite expectations that embedding based on \textit{Expert}-level descriptions would perform best when queried with CSPLib descriptions, this is not observed. Instead, the embeddings containing source code and \textit{Intermediate}-level descriptions achieve the highest MRR of $0.9051$. This is potentially due to CSPLib descriptions often being not truly high-level; for instance, they rarely incorporate extensive mathematical notation, whereas the generated \textit{Expert}-level descriptions make heavy use of it. In other words, CSPLib descriptions tend to be long and verbose rather than direct. A similar issue arises with \textit{Novice}-level descriptions, which include extensive explanations of fundamental concepts. One might also expect that the embeddings containing all description levels would perform best. However, a possible explanation for its under-performance is that incorporating \textit{Expert}-level descriptions dilutes the overall representation by making it more verbose and redundant, reducing its conciseness. 

We also entered CSPLib description queries for models not currently included in the CP-Model-Zoo database. 
Although this assessment is subjective, we found that the rankings produced by these queries were highly coherent and meaningful. The detailed results (using embedding SC+D2) of these queries are presented in the Fig.~\ref{anxD:other}.

\section{Web Application}
\label{sec:web}
To facilitate broader access to our research outcome by any interested party, a web-based application was developed as a demonstrative prototype, using the \texttt{gradio}\footnote{\url{https://www.gradio.app/}} Python framework. The application is hosted online\footnote{\url{https://cp-model-zoo.info.ucl.ac.be/}}
and uses as a knowledge base the SC+D2 embeddings set (see Section \ref{sec:exp}) for each combinatorial optimization problem available in it, 67 of these in total.

The end-user is able to interact with the tool by inputting a description of their problem in the textbox, found in the bottom left section of the application. Consecutively, the application retrieves and displays the ranked list of the top five most relevant CP models on the left-hand panel, above the textbox. Each model in this list is implemented as a clickable button, and upon selection, the corresponding MiniZinc source code is displayed on the right-side panel of the screen.

In Figure \ref{fig:gui}, we present a snapshot of the Graphical User Interface (GUI) of the application, upon execution of the following sample user query: 
\begin{verbatim}
"You need to put classes in a timetable so students don’t have two at the 
same time and don’t get overloaded in one semester."
\end{verbatim}

\begin{figure}[ht]
    \centering
    \includegraphics[width=0.8\linewidth]{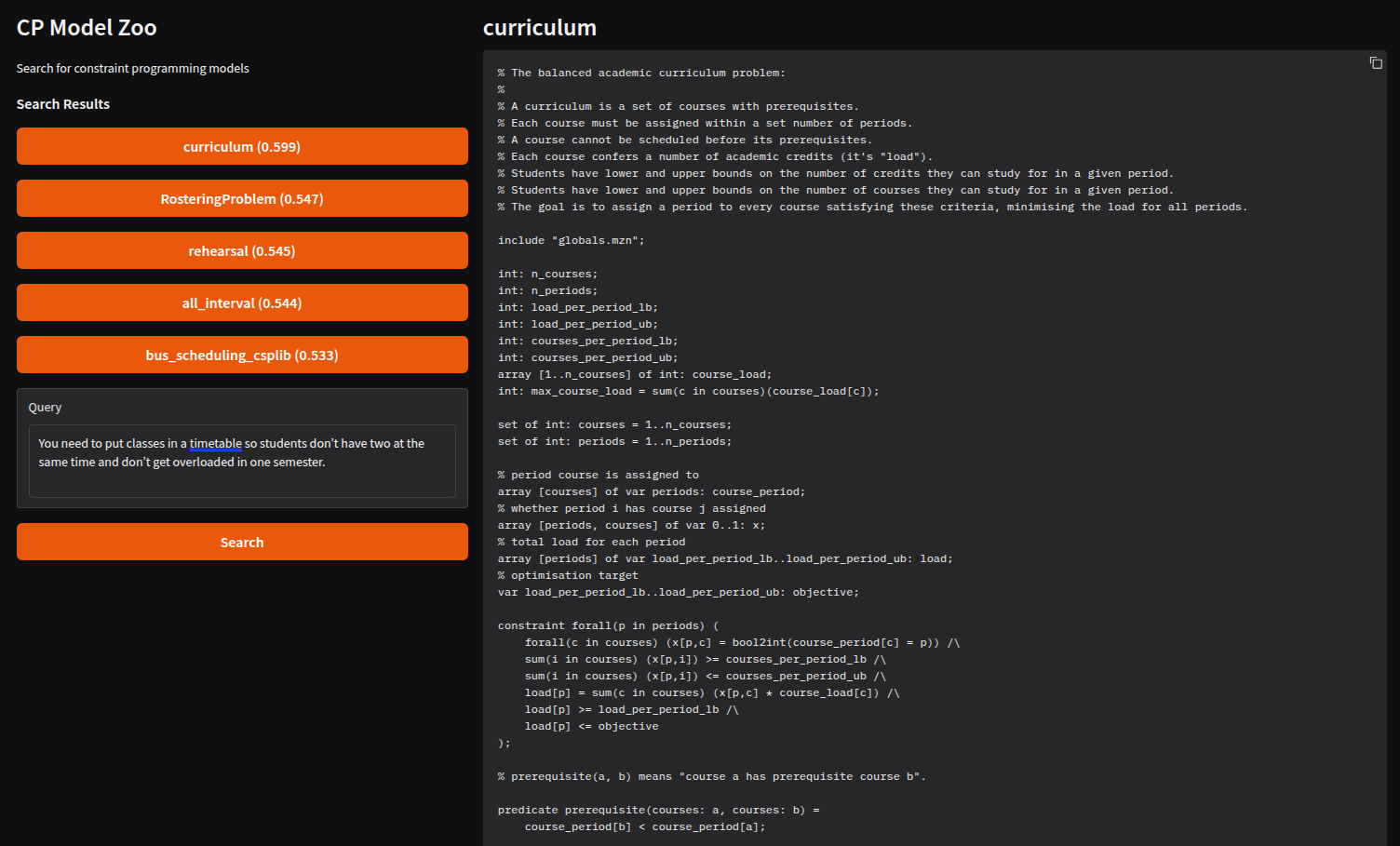}
    \caption{Snapshot of the GUI of the web application, developed to facilitate access to the tool.}
    \label{fig:gui}
\end{figure}

\section{Conclusion \& Future Work}
\label{sec:concl}

In our work, we introduced CP-Model-Zoo, a tool designed to retrieve the closest source code model from a database based on a user's natural language description of a combinatorial optimization problem. 
Our system aims to connect problem descriptions made by a possibly non-CP expert with expert-validated CP models by using high-quality models accumulated over the years by the community.

Our experimental results demonstrate that CP-Model-Zoo achieves excellent accuracy in retrieving relevant models, even when the queries are simulated at different levels of expertise. 
We believe this tool, which requires minimal human intervention, offers interesting value as an intelligent tutoring system for CP. It could be hosted on the CSPLib, and the continuous addition of new clean models to it could be a community effort.

\clearpage


\bibliography{LLM4CPbib}

\end{document}